\documentclass[sigconf]{acmart}
\usepackage[linesnumbered, ruled, vlined]{algorithm2e}
\usepackage{xspace}
\usepackage{amsmath}
\usepackage{balance}
\usepackage{color}
\usepackage{graphicx, float, overpic}
\usepackage{setspace}
\usepackage{enumitem}
\usepackage{multirow, bm}
\usepackage{textcomp}
\usepackage{subfigure}

\usepackage{algorithmic}

\usepackage{caption}
\renewcommand{\paragraph}[1]{
    \vskip 1mm
    \noindent
    {\bf #1.}
}

\theoremstyle{plain}

\theoremstyle{definition}

\AtBeginDocument{%
  }

\setcopyright{acmlicensed}
\copyrightyear{2025}
\acmYear{2025}
\acmDOI{10.1145/3746252.3761228}
\acmConference[CIKM'25]{Proceedings of the 34th ACM International Conference on Information and Knowledge Management}{ November 10--14, 2025}{Seoul, Republic of Korea.}
\acmBooktitle{Proceedings of the 34th ACM International Conference on Information and Knowledge Management (CIKM '25), November 10--14, 2025, Seoul, Republic of Korea}
\acmISBN{979-8-4007-2040-6/2025/11}
\acmDOI{10.1145/3746252.3761228}
\begin{document}

\title{Rethinking Client-oriented Federated Graph Learning}

\author{Zekai Chen}
\orcid{0009-0005-4591-9488}
\email{zackchen02@163.com}
\affiliation{%
    \institution{Beijing Institute of Technology}
    \city{Beijing}
    \country{China}
}

\author{Xunkai Li}
\orcid{0000-0002-1230-7603}
\email{cs.xunkai.li@gmail.com}
\affiliation{%
    \institution{Beijing Institute of Technology}
    \city{Beijing}
    \country{China}
}

\author{Yinlin Zhu}
\orcid{0009-0009-9181-6972}
\email{ylzhuawesome@163.com}
\affiliation{%
    \institution{Sun Yat-sen University}
    \city{Guangzhou}
    \state{Guangdong}
    \country{China}
}

\author{Rong-Hua Li}
\orcid{0000-0002-3105-5325}
\email{lironghuabit@126.com}
\affiliation{%
    \institution{Beijing Institute of Technology}
    \city{Beijing}
    \country{China}
}

\author{Guoren Wang}
\orcid{0000-0002-0181-8379}
\email{wanggrbit@gmail.com}
\affiliation{%
    \institution{Beijing Institute of Technology}
    \city{Beijing}
    \country{China}
}


\begin{abstract}

    As a new distributed graph learning paradigm, Federated Graph Learning (FGL) facilitates collaborative model training across local systems while preserving data privacy. 
    We review existing FGL approaches and categorize their optimization mechanisms into:
    (1) Server-Client (S-C), where clients upload local model parameters for server-side aggregation and global updates; 
    (2) Client-Client (C-C), which allows direct exchange of information between clients and customizing their local training process. 
    We reveal that C-C shows superior potential due to its refined communication structure. 
    However, existing C-C methods broadcast redundant node representations, incurring high communication costs and privacy risks at the node level. To this end, we propose FedC4, which combines graph \underline{C}ondensation with \underline{C}-\underline{C} \underline{C}ollaboration optimization. Specifically, FedC4 employs graph condensation technique to refine the knowledge of each client's graph into a few synthetic embeddings instead of transmitting node-level knowledge. Moreover, FedC4 introduces three novel modules that allow the source client to send distinct node representations tailored to the target client's graph properties. Experiments on eight public real-world datasets show that FedC4 outperforms state-of-the-art baselines in both task performance and communication cost. Our code is now available on https://github.com/Ereshkigal1/FedC4.

\end{abstract}

    

\maketitle
\section{Introduction}
\label{sec: intro}

   Graph Neural Networks (GNNs) have emerged as powerful tools for capturing the underlying topological and attribute patterns in graph-structured data, achieving notable success in real-world scenarios such as biomedical~\cite{qu2023app_gnn_bio2}, recommendation~\cite{recommend_system}, and finance~\cite{qiu2023app_gnn_fina3}. However, most existing GNN approaches rely on the assumption of centralized data storage, where the entire graph data is collected and accessed by a single institution. In contrast, in many real-world scenarios, graph data is inherently distributed across multiple institutions, with each institution only able to access its own private data due to privacy or storage constraints~\cite{FedSage,fedgta}. 
    This assumption-reality gap introduces significant challenges in enabling GNNs to effectively harness the full potential of collective intelligence in the decentralized setting.

    To this end, Federated Graph Learning (FGL) has emerged as a promising solution, combining the advantages of federated learning with GNNs. In this framework, multiple local graph systems collaborate to train a global GNN, with coordination managed by a trusted central server. Existing FGL optimization strategies can be classified into two paradigms:
    \textbf{(1) Server-Client (S-C)}, where clients upload local parameters for server-side aggregation and global updates. Approaches in this paradigm aim to enhance the global model's generalizability through modifications to training modules or architecture~\cite{GCFL,fedgta}.
    \textbf{(2) Client-Client (C-C)}, which facilitates the exchange of information like node embeddings between clients, enabling each client to customize its training process based on information from other clients. Research in this paradigm is dedicated to developing cross-client collaboration and local training processes for improved performance~\cite{FedSage,fedgcn}.

    \begin{figure*}[htbp]
    \centering
    \includegraphics[width=1.0\textwidth]{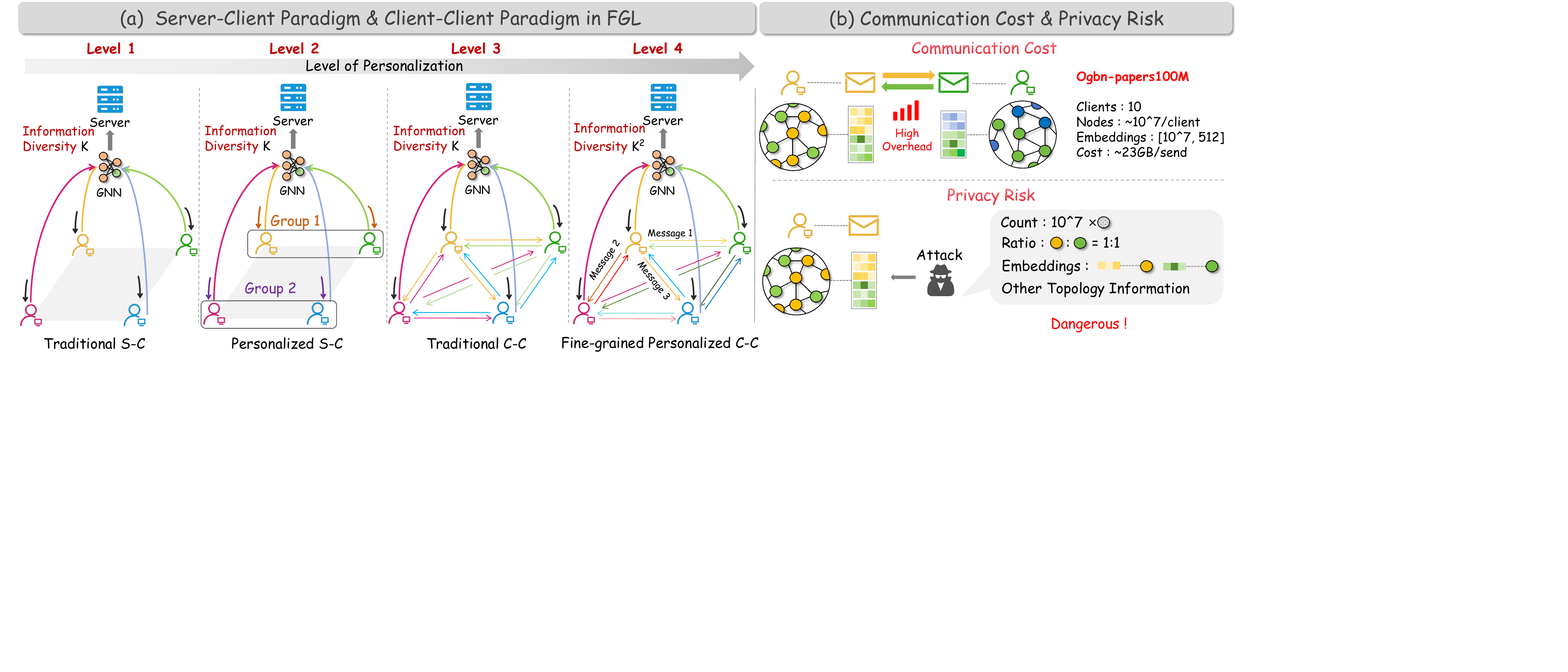}
    \caption{(a) \textbf{Progressive relationship in personalization potential in S-C and C-C:} \textit{Traditional S-C:} Clients send local updates to a central server for global aggregation. \textit{Personalized S-C:} The server clusters clients and provides group-specific aggregation. \textit{Traditional C-C:} Clients exchange identical information with all other clients. \textit{Fine-Grained Personalized C-C:} Clients exchange distinct and personalized information with different clients. (Different colored arrows represent unique information exchanged between clients.) (b) A key challenge is the substantial communication overhead and heightened risk of privacy leakage.}
    \label{fig:intro1}
    \end{figure*}
    
    Despite the effectiveness of both paradigms, we compare them in terms of information diversity to highlight their progressive relationship in personalization potential. As shown in Fig.~\ref{fig:intro1}(a), in the S-C paradigm, each of the $K$ clients uploads local information (e.g., model parameters) to a central server, which aggregates it into a global model in traditional S-C (\textbf{Level 1}) or up to $K$ distinct models in personalized S-C (\textbf{Level 2}). While this allows some personalization, the information diversity is limited to $K$ pieces, and does not support direct local optimization. In contrast, the broadcast-based C-C paradigm not only covers all of the S-C processes but also enables clients to share their local information, enhancing local optimization with global features and topology. Despite there being $K^2$ types of inter-client communication relations, this results in only $K$ distinct pieces of information (\textbf{Level 3}).

    Hence, the broadcast-based C-C overshadows its personalization potential due to the lack of a refined communication design. Intuitively, a more fine-grained personalized C-C approach allows each source client to tailor information for each target client instead of broadcasting the same information, increasing the diversity of inter-client communication to $K^2$ distinct pieces (\textbf{Level 4}). This motivates us to rethink the design of the FGL optimization strategy under the C-C paradigm to achieve fine-grained personalization.

    Moreover, to devise an effective FGL optimization strategy within the fine-grained personalized C-C paradigm, a critical challenge lies in addressing the substantial communication overhead and the heightened risk of privacy leakage, as illustrated in Fig.~\ref{fig:intro1}(b). First, all existing C-C technologies transmit data at the node level, meaning that as the volume of graph data grows, the communication cost becomes increasingly prohibitive. From a privacy perspective, traditional C-C methods such as FedSage+ \cite{FedSage} and FedGCN \cite{fedgcn} rely on transferring node hidden representations or propagated features. However, extensive studies have shown that these approaches inherently expose significant security vulnerabilities \cite{privacy_survey}. Although FedDEP \cite{feddep} enhances security with noiseless differential privacy, attackers can still infer sensitive information (e.g., node quantity) from side-channel data.
    
    Building on these insights, we propose a framework combining graph \underline{C}ondensation with \underline{C}-\underline{C} \underline{C}ollaboration optimization (FedC4). Specifically, FedC4 employs a graph condensation technique to refine the knowledge of each client's private graph into a few synthetic node embeddings instead of directly transmitting node-level knowledge, so as to achieve low-cost and high-privacy knowledge sharing among clients. Moreover, to achieve fine-grained personalized local optimization, FedC4 introduces three novel modules that allow the source client to send distinct node representations tailored to the target client's graph properties, thereby enhancing its local optimization with global features and topology. 

Our main contributions can be summarized as: (1) \textbf{New Observation.} This work introduces a novel viewpoint by categorizing existing FGL strategies based on personalization potential. (2) \textbf{New Paradigm.} We introduce a novel framework, \underline{Fed}erated Graph Learning with Graph \underline{C}ondensation and \underline{C}lient-\underline{C}lient \underline{C}ollaboration (\textbf{FedC4}), designed to enhance the C-C paradigm in FGL. By incorporating personalized and fine-grained inter-client knowledge sharing and local condensation, this framework effectively address challenges in C-C FGL. (3) \textbf{SOTA Performance.} We conduct comprehensive experiments on 8 datasets including transductive, inductive, heterogeneous, and large-scale graph settings. FedC4 outperforms the SOTA baselines with an average performance gain of 1.73\% and an efficiency improvement of up to 1000$\times$.


\section{Preliminaries and Related Work}
\subsection{Preliminaries}
\paragraph{Graph Neural Network}
GNNs are a class of deep learning models that operate on graph-structured data, leveraging both topology and node features to learn representations. A graph \( G = (A, X, Y) \) consists of an adjacency matrix \( A \), a node feature matrix \( X \), and node labels \( Y \). GNNs iteratively aggregate and update node features through a message-passing process. For example, the \( \ell \)-th layer of a GCN~\cite{GCN} propagates information as:
\begin{equation}
H^{(\ell)} = \text{ReLU}\left(\hat{A} H^{(\ell-1)} W^{(\ell)}\right),
\end{equation}
where \( H^{(0)} = X \), \( \hat{A} \) is the normalized adjacency matrix with self-loops, and \( W^{(\ell)} \) is the trainable weight matrix.

After \( L \) layers, GNNs capture \( L \)-hop neighborhood information, and the final node embeddings can be written as:
\begin{equation}
H = f(A, X),
\end{equation}
where \( f \) is the GNN model. The final embeddings \( h_i^{(L)} \) are passed to a classifier \( F \) for task-specific predictions:
\begin{equation}
z_i = F(h_i).
\end{equation}

\noindent 

\paragraph{Federated Graph Learning} FGL extends federated learning to graph-structured data, enabling clients to train local GNNs on private subgraphs while preserving data privacy. The optimization objective in FGL for a node classification task can be formulated as:
\begin{equation}
    \min_{\theta} \frac{1}{N} \sum_{i=1}^N \frac{|V^i|}{|V|} \mathcal{L}(f_{\theta}(V^i), Y^i),
\end{equation}
where \( V^i \) represents the set of nodes in the subgraph owned by client \( i \), \( Y^i \) are the corresponding labels, \( f_{\theta} \) denotes the GNN parameterized by \( \theta \), and \( \mathcal{L} \) is the task-specific loss.

In the FGL framework, each client independently trains a local GNN and then shares model parameters. The global model is then updated by aggregating these local updates in server-side, and then shared with all clients for the next round of training.

\paragraph{Graph Condensation} GC reduces large-scale graph data into a smaller, representative graph \( S = (A', X', Y') \), where \( A' \in \mathbb{R}^{N' \times N'} \) represents the adjacency matrix, \( X' \in \mathbb{R}^{N' \times d} \) denotes node features, and \( Y' \in \{1, \dots, C\}^{N'} \) indicates node labels. The objective is to train a GNN \( f_{\theta} \) on \( S \) with performance comparable to training on the original graph \( G\), while significantly reducing computational costs.

GC optimizes the synthesized graph \( S \) by minimizing the task-specific loss associated with the model's predictions:
\begin{equation}
\min_{S} \mathcal{L}\left(\text{classifier}(f_{\theta}(S)), Y\right),
\end{equation}
where \( \mathcal{L} \) is the loss function (e.g., cross-entropy). Synthesized graph \( S \) is designed to retain the structural and feature distributions of original graph \( G \), ensuring its utility in downstream tasks.

\subsection{Related Work}

\paragraph{Graph Neural Networks}
GNNs have advanced from early attempts to extend convolution to graphs~\cite{bruna2013spectral}, which often involved high parameter counts. GCN~\cite{GCN} addressed this using a first-order Chebyshev filter to capture local neighborhoods, while GAT~\cite{GAT} introduced attention mechanisms for weighted aggregation. GraphSAGE~\cite{GraphSAGE} further enhanced message aggregation with learnable functions. More research on GNNs can be found in surveys~\cite{gnnsurvey1,gnnsurvey2}.

\paragraph{Federated Graph Learning}
FGL combines the principles of federated learning with GNNs and has emerged as a prominent research direction in recent years. The S-C paradigm, exemplified by methods such as FedLIT~\cite{FedLIT} and FedGTA~\cite{fedgta}, aggregates client parameters on a central server. In contrast, the C-C paradigm enables direct message exchanges between clients. Representative methods include FedSage+~\cite{FedSage}, which facilitates subgraph training via embedding exchanges, FedDEP~\cite{feddep}, which incorporates dependency-aware communication, and FedGCN~\cite{fedgcn}, which exchanges compressed intermediate features during GNN training. However, these methods still face challenges in fine-grained personalization, communication costs, and privacy risk, highlighting the need for more effective solutions.

\paragraph{Graph Condensation}
GC aims to efficiently distill a small-scale synthetic graph from a large-scale graph while ensuring that models trained on the condensed graph achieve performance comparable to those trained on the original graph. GCond~\cite{GCOND} leverages gradient matching to optimize adjacency matrices and node features, while SGDD~\cite{SGDD} incorporates Laplacian energy matching to enhance structural representation. MCond~\cite{MCond} focuses on feature condensation for multimodal graph data, and SFGC~\cite{SFGC} introduces a global consistency mechanism to improve condensation performance.

\section{Proposed Framework: FedC4}

\begin{figure*}[htbp]
    \centering
    \includegraphics[width=1.0\textwidth]{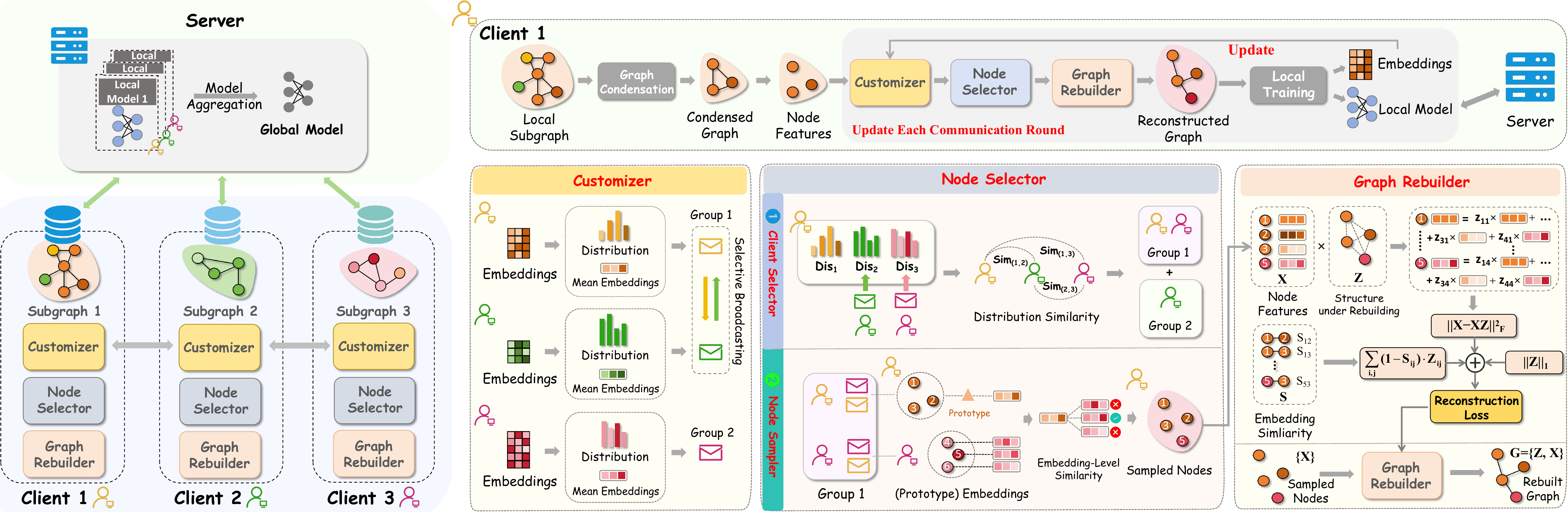}
    \caption{Overview of our proposed FedC4 framework. The left side illustrates the overall workflow of server-side broadcasting and aggregation. The right side illustrates client-side operations including local condensation and three core modules.}
    \label{fig:framework}
\end{figure*}

\subsection{Framework Overview} 

As illustrated in Fig.~\ref{fig:framework}, FedC4 consists of four key components: Local Graph Condensation, Customizer Module (CM), Node Selector Module (NS), and Graph Rebuilder Module (GR). While the figure describes the high-level process of the framework, the detailed design and functionalities of the GNN models, as well as the individual modules, will be elaborated in subsequent sections.

We also notice a recent work, FedGC~\cite{fedgc}, which integrates FGL concepts into the GC process by decoupling the gradient matching procedure: clients compute local gradients, and the server aggregates them to optimize a global condensed graph. Essentially, FedGC applies FGL techniques to support the task of GC.

In contrast, FedC4 leverages GC techniques to enhance C-C FGL. It targets fine-grained personalization, communication efficiency, and privacy protection under the C-C paradigm. These goals are achieved by combining GC with three dedicated modules: CM, NS, and GR. This design makes FedC4 fundamentally different from FedGC in both objective and architecture: FedGC uses FL to facilitate GC, whereas FedC4 uses GC to empower C-C FGL.

\subsection{Local Graph Condensation}

As mentioned in Sec.~\ref{sec: intro}, the C-C paradigm introduces significant communication costs and privacy challenges. GC addresses these issues by synthesizing compact and secure graphs \(S\) that retain the structural and semantic properties of the original graph \(\mathcal{G}\). 

GC initialization begins by randomly initializing the feature matrix \(\mathbf{X}'\) from a Gaussian distribution. Labels \(\mathbf{Y}'\) are selected to match the label distribution of \(\mathcal{G}\). The adjacency matrix \(\mathbf{A}'\) is dynamically generated by passing \(\mathbf{X}'\) through a randomly initialized trainable MLP, allowing the condensed graph to adaptively learn its structure during optimization.

To ensure that the condensed graph preserves the original graph's properties, the optimization process minimizes the gradient matching loss of the GNN parameterized by \(\theta\):
\begin{equation}
\mathcal{L}_{\text{mat}} = \sum_{i=1}^{|\theta|} \|\nabla_{\theta_i} \mathcal{L}^G - \nabla_{\theta_i} \mathcal{L}^S\|^2 ,
\label{eq:gradient_match}
\end{equation}
where \(\mathcal{L}^G\) and \(\mathcal{L}^S\) denote the cross-entropy losses of the original and condensed graphs, respectively. $\mathcal{L}_{\text{mat}}$ is then used to optimize the MLP. To further refine \(\mathbf{A}'\), a sparsification step is applied by thresholding edge weights:
\begin{equation}
\mathbf{A}'_{ij} = 
\begin{cases} 
\mathbf{A}'_{ij}, & \text{if } \mathbf{A}'_{ij} > \delta, \\
0, & \text{otherwise},
\end{cases}
\label{eq:sparsification}
\end{equation}

\noindent where \(\delta\) is the sparsification threshold, consistent with same hyperparameter selection described in GCond~\cite{GCOND}.
The complete process of local condensation is also described in GCond~\cite{GCOND}.

\subsection{Customizer (CM)}

In the C-C paradigm, the CM module addresses the limitation of broadcasting identical content by facilitating collaboration through the selective sharing of local embedding statistics, such as distributions and prototype embeddings. During the first round, local statistics are shared with all clients to establish a baseline understanding. In subsequent rounds, these statistics are selectively broadcast to same-cluster clients (\(\mathcal{C}_{\text{same}}\)), as determined by the clustering results from the Node Selector module in the previous round, enabling personalized and selective collaboration.

For a client \(c\), local node embeddings \(\mathbf{H}_c = \{ \mathbf{h}_i \mid i \in \mathcal{V}_c \}\) and node number \(\mathcal{V}_c\) are used to compute the embedding distribution \(\text{Dis}_c\) and prototype embedding \(\mu_c\):
\begin{equation}
\text{Dis}_c = \{ \|\mathbf{h}_i\| \mid i \in \mathcal{V}_c \}, \quad 
\mu_c = \frac{1}{|\mathcal{V}_c|} \sum_{i \in \mathcal{V}_c} \mathbf{h}_i,
\label{eq:cm8}
\end{equation}
where \(\|\mathbf{h}_i\|\) is the norm of embedding \(\mathbf{h}_i\). These statistics are normalized for consistency across clients and then broadcast according to the communication strategy:

\begin{equation}
\mu_{\text{global}} = \frac{1}{C} \sum_{c=1}^C \mu_c, \sigma_{\text{global}} = \sqrt{\frac{1}{C} \sum_{c=1}^C \|\mu_c - \mu_{\text{global}}\|^2}.
\label{eq:cm9}
\end{equation}

To ensure consistent scale and distribution of embeddings across all clients, we normalize the individual client statistics using the global metrics as follows:

\begin{equation}
\mu_c' = \frac{\mu_c - \mu_{\text{global}}}{\sigma_{\text{global}} + \epsilon},
\text{Dis}_c' = \left\{ \frac{\|\mathbf{h}_i\| - \mu_{\text{dis}}}{\sigma_{\text{dis}}} \mid \|\mathbf{h}_i\| \in \text{Dis}_c \right\},
\label{eq:cm10}
\end{equation}

where \(\epsilon\) is a small constant to prevent division by zero, and \(\mu_{\text{dis}}\) and \(\sigma_{\text{dis}}\) are the mean and standard deviation of the norms of embeddings across all clients, respectively. The updated values \(\text{Dis}_{c'}\) and \(\mu_{c'}\) are then selectively broadcast to clients based on clustering results 
\(\mathcal{C}_{\text{target}}\) from the Node Selector, enabling fine-grained and personalized collaboration:
\begin{equation}
\{\text{Dis}_{c'}, \mu_{c'} \mid c' \in \mathcal{C}_{\text{target} \setminus \{c\}} \},
\label{eq:cm11}
\end{equation}
where \(\mathcal{C}_{\text{target}}\) refers to all clients (\(\mathcal{C}\)) in the first round and same-cluster clients (\(\mathcal{C}_{\text{same}}\)) in subsequent rounds.

By equipping clients with normalized global statistics and leveraging selective communication based on previous clustering results, the CM module ensures informed local decision-making, improving fine-grained collaboration in the C-C paradigm. The complete process of CM module is described in Algorithm ~\ref{alg:broadcaster} in Appendix.

\subsection{Node Selector (NS)}

In addition to selective broadcasting, it is essential to tailor the messages for each client. By identifying representative nodes that capture key information from local graphs, NS ensures that exchanged data is both relevant and contextually aware. This targeted strategy addresses the limitations of uniform broadcasting, enabling more personalized and efficient collaboration.

For each client $c \in \mathcal{C}$, the module computes the Sliced Wasserstein Distance (SWD) between its distribution $\text{Dis}_c$ and those of other clients $\text{Dis}_{c'}$, where $c' \in \mathcal{C} \setminus \{c\}$:
\begin{equation}
\text{SWD}_{c,c'} \! = \!\!\! \int_{S^{d-1}} \!\!\sup_{\theta  \in S^{d-1}} \! \left| \! \int_{\mathbb{R}^d}\!\!\!\! \theta \cdot x \, d\text{Dis}_c(x) \! - \!\!\! \int_{\mathbb{R}^d} \!\!\!\!\theta \cdot y \, d\text{Dis}_{c'}(y)  \right | d\theta,
\label{SWD}
\end{equation}
where $S^{d-1}$ represents the unit sphere in $\mathbb{R}^d$. SWD effectively reduces the computational complexity in high-dimensional spaces compared to the traditional WD by projecting the distributions onto a line and measuring their one-dimensional discrepancies.

Clients are then grouped into clusters $\{\mathcal{C}_1, \mathcal{C}_2, \dots\}$ based on SWD similarity. Within each cluster $\mathcal{C}_c$, the cosine similarity between client $c$’s node embeddings $\mathbf{h}_i$ and prototype embedding $\mu_{c'}$ of other clients $c' in \mathcal{C}_c$ is computed:
\begin{equation}
S(\mathbf{h}_i, \mu_{c'}) = \frac{\mathbf{h}_i \cdot \mu_{c'}}{\|\mathbf{h}_i\| \|\mu_{c'}\|}.
\label{cossim}
\end{equation}
Nodes exhibiting similarity scores exceeding a designated threshold \(\tau\) are incorporated into the representative set \(\mathcal{S}_c\). The selection criteria for \(\tau\) and its consequent effects on the accuracy curve are comprehensively elucidated in hyper-parameter study in Sec.~\ref{par:hyper}.

By selecting nodes aligned with shared semantic patterns, the NS module identifies and selects representative nodes for effective and fine-grained collaboration. The complete process of NS module is described in Algorithm ~\ref{alg:nodeselector} in Appendix.

\subsection{Graph Rebuilder (GR)}

The GR module generates appropriate structures for all candidate nodes selected by NS by leveraging the concept of self-expressive graph reconstruction~\cite{GC}, rather than simply linking newly introduced nodes to the original condensed graph. To adapt to the FGL settings, GR incorporates both embedding similarity loss and sparsification loss, enabling the generation of graph structures that are well-suited for distributed learning. Specifically, GR constructs adaptive and reliable graph topologies for the candidate nodes selected by NS, ensuring that the reconstructed graph effectively preserves essential structural and semantic information while maintaining sparsity for efficiency. By constructing adaptive graph topologies tailored to the specific features and relationships of the candidate nodes, GR ensures fine-grained personalized structual information.

For each client \(c \in \mathcal{C}\), GR takes the node feature matrix \(\mathbf{X}_c\) of the sampled nodes as an extra input. Embedding similarity between 
candidate nodes can be computed as:
\begin{equation}
S_{ij} = \frac{\mathbf{h}_i \cdot \mathbf{h}_j}{\|\mathbf{h}_i\| \|\mathbf{h}_j\|},
\label{embsim}
\end{equation}
where \(S_{ij}\) captures the similarity between nodes \(i\) and \(j\). Using this similarity, the module updates the reconstructed adjacency matrix \(\mathbf{Z}\) by minimizing the reconstruction loss:
\begin{equation}
\mathcal{L}_{\text{Rec}} =
\alpha \|\mathbf{X}_c - \mathbf{X}_c \mathbf{Z}\|_F^2 
+ \beta \|\mathbf{Z}\|_1 
+ \sum_{i,j} (1 - S_{ij}) \cdot \mathbf{Z}'_{ij},
\label{lossrec}
\end{equation}

The reconstruction loss \(\mathcal{L}_{\text{Rec}}\) consists of three terms: the first term, \(\|\mathbf{X}_c - \mathbf{X}_c \mathbf{Z}\|_F^2\), ensures that the reconstructed adjacency matrix \(\mathbf{Z}\) preserves the structural relationships of the original graph in the feature space. The second term, \(\|\mathbf{Z}\|_1\), promotes sparsity in \(\mathbf{Z}\), retaining only the most significant edges and avoiding overly dense graphs. The third term, \(\sum_{i,j} (1 - S_{ij}) \cdot \mathbf{Z}'_{ij}\), aligns the reconstructed graph with node similarity \(S_{ij}\), penalizing edges inconsistent with the semantic relationships captured by the embeddings. The hyperparameters \(\alpha\) and \(\beta\) control the balance between these terms.

The reconstructed graph is then used to train local GNN, with the updated model uploaded for global aggregation. Although partially dependent on the smoothing effects of graph neural networks, the excellent performance on heterogeneous graph datasets in Sec.~\ref{par:performance} demonstrates that the GR module has robust capabilities for handling heterogeneous graphs. The detailed process of GR module is described in Algorithm~\ref{alg:graph_rebuilder} in Appendix.

\section{Theoretical Analysis}
\subsection{Privacy Protection}
\label{sec: privacy}

In this section, we theoretically demonstrate that GC inherently provides privacy protection when transmitting synthetic graph embeddings in C-C paradigms.

Before presenting the proof, we clarify the privacy scenario GC aims to protect against. The objective is to minimize the risk of inferring individual node attributes from synthetic graph embeddings. This is particularly relevant in scenarios where node data may contain sensitive information. The privacy analysis presented here focuses on ensuring that changes in individual nodes of the original graph \( \mathcal{G} \)  do not significantly influence the derived synthetic graph \( \mathcal{S} \), thus maintaining a high level of privacy.

\paragraph{Proof}  
Given the synthetic graph \( \mathcal{S} = (\mathcal{V}_{\text{syn}}, \mathcal{E}_{\text{syn}}, \mathbf{X}_{\text{syn}}) \) and the original graph \( \mathcal{G} = (\mathcal{V}, \mathcal{E}, \mathbf{X}) \) with \( n = |\mathcal{V}| \) nodes, the training loss is expressed as:
\begin{equation}
\mathcal{L}(\mathcal{S}, \mathcal{G}) = \frac{1}{n} \sum_{i=1}^{n} \ell(f_{\theta}(\mathbf{x}_i), y_i),
\end{equation}
where \( \ell \) is the sample-wise loss, \( f_{\theta} \) is the model parameterized by \( \theta \), and \((\mathbf{x}_i, y_i)\) denotes the feature-label pair. Removing node \( v_j \) modifies the graph to \( \mathcal{G}^{-j} \), the adjusted loss is:
\begin{equation}
\mathcal{L}(\mathcal{S}, \mathcal{G}^{-j}) = \frac{1}{n-1} \sum_{\substack{i=1 \\ i \neq j}}^{n} \ell(f_{\theta}(\mathbf{x}_i), y_i).
\end{equation}

\noindent The gradient difference due to node removal expands to:
\begin{align}
\Delta \nabla &= \nabla \mathcal{L}(\mathcal{S}, \mathcal{G}) - \nabla \mathcal{L}(\mathcal{S}, \mathcal{G}^{-j})  \\
&= \frac{1}{n} \nabla \ell(f_{\theta}(\mathbf{x}_j), y_j) - \frac{1}{n(n-1)} \sum_{\substack{i=1 \\ i \neq j}}^{n} \nabla \ell(f_{\theta}(\mathbf{x}_i), y_i). \notag
\end{align}

\noindent Assuming the gradient of each sample is bounded as \( \|\nabla \ell(f_{\theta}(\mathbf{x}_i), y_i)\| \leq C \), gradient difference is bounded as:
\begin{equation}
\|\Delta \nabla\| \leq \frac{C}{n} + O\left(\frac{1}{n^2}\right).
\end{equation}
For large \( n \), the second term is negligible, yielding:
\begin{equation}
\|\Delta \nabla\| \leq O\left(\frac{1}{n}\right).
\end{equation}

\noindent This bounded gradient difference propagates through the GNN layers, impacting the node embeddings as:
\begin{equation}
\Delta_j = \|\mathbf{H}_{\text{syn}} - \mathbf{H}_{\text{syn}}^{-j}\| \leq \|\mathbf{H}^{-1}\| \cdot \|\Delta \nabla\|,
\end{equation}
where \(\|\mathbf{H}^{-1}\|\) is the norm of the inverse of the Hessian matrix. Substituting the bound on \( \Delta \nabla \), we obtain:
\begin{equation}
\Delta_j \leq O\left(\frac{m}{n}\right),
\end{equation}
where \( m = |\mathcal{V}_{\text{syn}}| \) is the size of the synthetic graph. \qed

\begin{table*}[t]
\centering
\caption{Performance comparison across datasets with varying compression ratios and baselines.}
\resizebox{\textwidth}{!}{
\begin{tabular}{c|c|c|c c c c c c c c c c c c c}
\specialrule{1.5pt}{1.5pt}{1.5pt}
\multirow{2}{*}{\textbf{Property}} & \multirow{2}{*}{\textbf{Dataset}} & \multirow{2}{*}{\textbf{Ratio}} 
& \multicolumn{2}{c}{\textbf{FL}}
& \multicolumn{3}{c}{\textbf{FL+Graph Reduction}} 
& \multicolumn{3}{c}{\textbf{FL+GC}}
& \multicolumn{3}{c}{\textbf{FGL C-C}}
& \multicolumn{1}{c}{\textbf{FGL S-C}}
& \multicolumn{1}{c}{\textbf{Ours}} \\ 
\cmidrule(lr){4-5} \cmidrule(lr){6-8} \cmidrule(lr){9-11} \cmidrule(lr){12-14} \cmidrule(lr){15-15} \cmidrule(lr){16-16} 
& & & FedAvg & FedDC & Random & Herding & Coarsening & GCond 
& DosCond & SFGC & FedSage+ & FedGCN & FedDEP & Best S-C & FedC4 \\ 
\midrule

\multirow{6}{*}{\textbf{Small}} 
&      & 8\%  &  &  & $81.35_{\scriptstyle \pm 0.33}$ & $81.87_{\scriptstyle \pm 0.25}$ & $82.63_{\scriptstyle \pm 0.56}$ & $83.61_{\scriptstyle \pm 1.14}$ & $82.24_{\scriptstyle \pm 0.41}$ & $83.58_{\scriptstyle \pm 0.59}$ &  &  &  &  & \cellcolor[HTML]{D5FDF9} \textbf{\boldmath $86.02_{\scriptstyle \pm 0.56}$} \\ 
& Cora & 4\%  & $84.70_{\scriptstyle \pm 0.61}$ & $84.79_{\scriptstyle \pm 0.69}$ & $80.61_{\scriptstyle \pm 1.04}$ & $80.63_{\scriptstyle \pm 0.96}$ & $81.81_{\scriptstyle \pm 0.62}$ & $82.97_{\scriptstyle \pm 0.85}$ & $81.83_{\scriptstyle \pm 0.49}$ & $83.15_{\scriptstyle \pm 0.77}$ & $83.69_{\scriptstyle \pm 0.73}$ & $84.32_{\scriptstyle \pm 0.55}$ & \cellcolor[HTML]{E9F8FD} $\underline{85.01_{\scriptstyle \pm 0.66}}$ & $84.97_{\scriptstyle \pm 0.70}$ & $84.98_{\scriptstyle \pm 0.65}$ \\  
&      & 2\%  &  &  & $79.03_{\scriptstyle \pm 1.25}$ & $79.74_{\scriptstyle \pm 1.11}$ & $80.73_{\scriptstyle \pm 0.70}$ & $81.36_{\scriptstyle \pm 1.33}$ & $81.49_{\scriptstyle \pm 0.62}$ & $81.42_{\scriptstyle \pm 0.98}$ &  &  &  &  & $84.33_{\scriptstyle \pm 0.80}$ \\  
\cmidrule{2-16}
&          & 8\% &  &  & $65.97_{\scriptstyle \pm 0.27}$ & $68.60_{\scriptstyle \pm 0.30}$ & $65.57_{\scriptstyle \pm 0.19}$ & $72.03_{\scriptstyle \pm 1.02}$ & $72.58_{\scriptstyle \pm 0.65}$ & $73.11_{\scriptstyle \pm 0.25}$ &  &  &  &  & \cellcolor[HTML]{D5FDF9} \textbf{\boldmath $75.95_{\scriptstyle \pm 0.66}$} \\  
& Citeseer & 4\% & $72.11_{\scriptstyle \pm 0.73}$ & $71.94_{\scriptstyle \pm 0.52}$ & $65.89_{\scriptstyle \pm 0.99}$ & $67.52_{\scriptstyle \pm 0.87}$ & $66.83_{\scriptstyle \pm 0.40}$ & $67.27_{\scriptstyle \pm 1.35}$ & $68.02_{\scriptstyle \pm 0.81}$ & $71.00_{\scriptstyle \pm 0.29}$ & $70.77_{\scriptstyle \pm 0.88}$ &  $74.37_{\scriptstyle \pm 0.49}$ & $73.83_{\scriptstyle \pm 0.86}$ & \cellcolor[HTML]{E9F8FD} $\underline{74.89_{\scriptstyle \pm 0.93}}$ & $74.84{\scriptstyle \pm 0.74}$ \\  
&          & 2\% &  &  & $65.73_{\scriptstyle \pm 1.56}$ & $65.73_{\scriptstyle \pm 1.02}$ & $67.91_{\scriptstyle \pm 0.45}$ & $67.46_{\scriptstyle \pm 2.24}$ & $70.41_{\scriptstyle \pm 1.22}$ & $69.23_{\scriptstyle \pm 0.66}$ &  &  &  &  & $73.91_{\scriptstyle \pm 1.17}$ \\  
\midrule

\multirow{6}{*}{\textbf{Medium}} 
&             & 0.4\% &  &  & $58.91_{\scriptstyle \pm 0.82}$ & $56.63_{\scriptstyle \pm 0.77}$ & $59.96_{\scriptstyle \pm 0.16}$ & $65.98_{\scriptstyle \pm 0.44}$ & $63.41_{\scriptstyle \pm 0.23}$ & $66.17_{\scriptstyle \pm 0.52}$ &  &  &  &  & \cellcolor[HTML]{D5FDF9} \textbf{\boldmath $67.66_{\scriptstyle \pm 0.38}$} \\  
& Arxiv & 0.2\% & $62.16_{\scriptstyle \pm 0.53}$ & $61.20_{\scriptstyle \pm 0.66}$ & $57.60_{\scriptstyle \pm 1.03}$ & $56.22_{\scriptstyle \pm 0.94}$ & $59.11_{\scriptstyle \pm 0.29}$ & $65.38_{\scriptstyle \pm 0.82}$ & $63.25_{\scriptstyle \pm 0.37}$ & $65.62_{\scriptstyle \pm 0.68}$ & $57.07_{\scriptstyle \pm 0.55}$ & $64.17_{\scriptstyle \pm 0.69}$ & $64.31_{\scriptstyle \pm 0.79}$ & $63.14_{\scriptstyle \pm 0.62}$ & \cellcolor[HTML]{E9F8FD} $\underline{66.95_{\scriptstyle \pm 0.76}}$ \\  
&            & 0.1\% &  &  & $57.29_{\scriptstyle \pm 1.89}$ & $55.98_{\scriptstyle \pm 1.22}$ & $57.38_{\scriptstyle \pm 0.41}$ & $64.81_{\scriptstyle \pm 1.17}$ & $62.33_{\scriptstyle \pm 0.52}$ & $65.09_{\scriptstyle \pm 0.57}$ &  &  &  &  & $65.42_{\scriptstyle \pm 1.36}$ \\  
\cmidrule{2-16}
&         & 1.4\% &  &  & $92.25_{\scriptstyle \pm 0.45}$ & $92.72_{\scriptstyle \pm 0.51}$ & $90.24_{\scriptstyle \pm 0.28}$ & $92.88_{\scriptstyle \pm 0.69}$ & $91.56_{\scriptstyle \pm 0.38}$ & $92.97_{\scriptstyle \pm 0.55}$ &  &  &  &  & \cellcolor[HTML]{D5FDF9} \textbf{\boldmath $94.61_{\scriptstyle \pm 0.72}$}  \\  
& Physics & 0.7\% & $92.74_{\scriptstyle \pm 0.43}$ & $94.02_{\scriptstyle \pm 0.37}$ & $91.37_{\scriptstyle \pm 0.75}$ & $92.37_{\scriptstyle \pm 0.33}$ & $89.73_{\scriptstyle \pm 0.22}$ & $92.13_{\scriptstyle \pm 1.01}$ & $90.91_{\scriptstyle \pm 0.42}$ & $92.41_{\scriptstyle \pm 0.63}$ & $93.01_{\scriptstyle \pm 0.68}$ & $92.88_{\scriptstyle \pm 0.65}$ & $93.72_{\scriptstyle \pm 0.80}$ & \cellcolor[HTML]{E9F8FD} $\underline{94.12_{\scriptstyle \pm 0.53}}$& $93.65_{\scriptstyle \pm 0.82}$ \\  
&         & 0.35\% &  &  & $91.55_{\scriptstyle \pm 1.63}$ & $91.53_{\scriptstyle \pm 0.82}$ & $88.42_{\scriptstyle \pm 0.40}$ & $91.48_{\scriptstyle \pm 1.64}$ & $90.52_{\scriptstyle \pm 0.68}$ & $91.86_{\scriptstyle \pm 0.79}$ &  &  &  &  & $93.41_{\scriptstyle \pm 1.33}$ \\  
\midrule

\multirow{6}{*}{\textbf{Inductive}} 
&        & 1\%   &  &  & $45.21_{\scriptstyle \pm 0.15}$ & $44.80_{\scriptstyle \pm 0.18}$ & $46.39_{\scriptstyle \pm 0.09}$ & $47.11_{\scriptstyle \pm 0.22}$ & $45.54_{\scriptstyle \pm 0.38}$ & $47.03_{\scriptstyle \pm 0.25}$ &       &    &   &       & \cellcolor[HTML]{D5FDF9} \textbf{\boldmath $48.62_{\scriptstyle \pm 0.27}$} \\  
& Flickr & 0.5\% & $46.09_{\scriptstyle \pm 0.23}$ & $43.73_{\scriptstyle \pm 0.21}$ & $44.75_{\scriptstyle \pm 0.31}$ & $44.46_{\scriptstyle \pm 0.47}$ & $46.14_{\scriptstyle \pm 0.13}$ & $47.06_{\scriptstyle \pm 0.28}$ & $45.01_{\scriptstyle \pm 0.78}$ & $46.86_{\scriptstyle \pm 0.42}$ & $45.27_{\scriptstyle \pm 0.38}$ & $46.44_{\scriptstyle \pm 0.52}$ & $47.36_{\scriptstyle \pm 0.44}$ & $48.11_{\scriptstyle \pm 0.30}$ & \cellcolor[HTML]{E9F8FD} $\underline{48.35_{\scriptstyle \pm 0.22}}$ \\  
&        & 0.1\% &  &  & $44.39_{\scriptstyle \pm 0.42}$ & $44.08_{\scriptstyle \pm 0.69}$ & $45.82_{\scriptstyle \pm 0.22}$ & $46.75_{\scriptstyle \pm 0.39}$ & $44.86_{\scriptstyle \pm 1.02}$ & $46.42_{\scriptstyle \pm 0.70}$ &       &   &    &       & $47.89_{\scriptstyle \pm 0.31}$ \\ 
\cmidrule{2-16}
&        & 1\%   &  &  & $87.82_{\scriptstyle \pm 0.73}$ & $89.37_{\scriptstyle \pm 0.86}$ & $88.12_{\scriptstyle \pm 0.66}$ & $91.55_{\scriptstyle \pm 0.49}$ & $91.39_{\scriptstyle \pm 0.33}$ & $91.44_{\scriptstyle \pm 0.38}$ &       &    &   &       & \cellcolor[HTML]{D5FDF9} \textbf{\boldmath $92.38_{\scriptstyle \pm 0.21}$} \\  
& Reddit & 0.5\% & $91.80_{\scriptstyle \pm 0.42}$ & \cellcolor[HTML]{E9F8FD} $\underline{91.85_{\scriptstyle \pm 0.38}}$ & $87.57_{\scriptstyle \pm 0.66}$ & $89.01_{\scriptstyle \pm 1.03}$ & $87.88_{\scriptstyle \pm 0.91}$ & $91.43_{\scriptstyle \pm 0.52}$ & $91.28_{\scriptstyle \pm 0.25}$ & $91.32_{\scriptstyle \pm 0.42}$ & OOT   & $91.84_{\scriptstyle \pm 0.55}$  & OOT & $91.80_{\scriptstyle \pm 0.33}$ & $91.73_{\scriptstyle \pm 0.38}$ \\  
&        & 0.1\% &  &  & $86.89_{\scriptstyle \pm 1.22}$ & $88.34_{\scriptstyle \pm 1.83}$ & $87.27_{\scriptstyle \pm 1.14}$ & $91.07_{\scriptstyle \pm 0.87}$ & $91.03_{\scriptstyle \pm 0.38}$ & $91.10_{\scriptstyle \pm 0.53}$ &       &   &    &       & $91.34_{\scriptstyle \pm 0.56}$ \\ 

\midrule

\multirow{3}{*}{\textbf{Large}} &                    & 0.4\% &  &  & $82.65_{\scriptstyle \pm 0.55}$ & $82.41_{\scriptstyle \pm 0.21}$ & $82.00_{\scriptstyle \pm 0.18}$ & $83.39_{\scriptstyle \pm 0.42}$ & $82.52_{\scriptstyle \pm 0.33}$ & $83.02_{\scriptstyle \pm 0.27}$ &  &  &  &  & \cellcolor[HTML]{D5FDF9} \textbf{\boldmath $85.35_{\scriptstyle \pm 0.45}$} \\  
& Products & 0.2\% & $82.31_{\scriptstyle \pm 0.47}$ & $78.41_{\scriptstyle \pm 0.38}$ & $82.38_{\scriptstyle \pm 0.71}$ & $81.99_{\scriptstyle \pm 0.39}$ & $81.72_{\scriptstyle \pm 0.32}$ & $83.08_{\scriptstyle \pm 0.85}$ & $82.33_{\scriptstyle \pm 0.47}$ & $82.79_{\scriptstyle \pm 0.56}$ & OOM & $84.20_{\scriptstyle \pm 0.53}$ & OOM & OOM & \cellcolor[HTML]{E9F8FD} $\underline{84.53_{\scriptstyle \pm 0.72}}$ \\  
&                  & 0.1\% &  &  & $81.77_{\scriptstyle \pm 1.31}$ & $82.30_{\scriptstyle \pm 0.60}$ & $81.45_{\scriptstyle \pm 0.48}$ & $82.81_{\scriptstyle \pm 1.42}$ & $81.94_{\scriptstyle \pm 0.73}$ & $82.57_{\scriptstyle \pm 0.80}$ &  &  &   &  & $84.17_{\scriptstyle \pm 0.99}$ \\  
\midrule

\multirow{3}{*}{\textbf{Hetero.}} &                & 1.4\% &  &  & $65.26_{\scriptstyle \pm 0.61}$ & $62.11_{\scriptstyle \pm 0.18}$ & $61.76_{\scriptstyle \pm 0.37}$ & $65.80_{\scriptstyle \pm 0.38}$ & $68.73_{\scriptstyle \pm 0.25}$ & $68.97_{\scriptstyle \pm 0.16}$ &  &  &  &  & \cellcolor[HTML]{D5FDF9} \textbf{\boldmath $71.05_{\scriptstyle \pm 0.27}$} \\  
& Empire & 0.7\% & $45.24_{\scriptstyle \pm 0.32}$ & $39.45_{\scriptstyle \pm 0.28}$ & $65.01_{\scriptstyle \pm 0.45}$ & $61.15_{\scriptstyle \pm 0.38}$ & $61.09_{\scriptstyle \pm 0.22}$ & $65.37_{\scriptstyle \pm 0.42}$ & $66.17_{\scriptstyle \pm 0.33}$ & $66.27_{\scriptstyle \pm 0.18}$ & \cellcolor[HTML]{E9F8FD} $\underline{69.68_{\scriptstyle \pm 0.70}}$ & $68.82_{\scriptstyle \pm 0.61}$ & $67.55_{\scriptstyle \pm 0.53}$ & $49.08_{\scriptstyle \pm 0.57}$ & $68.19_{\scriptstyle \pm 0.44}$ \\  
&                      & 0.35\% &  &  & $64.90_{\scriptstyle \pm 1.08}$ & $60.54_{\scriptstyle \pm 0.75}$ & $60.14_{\scriptstyle \pm 0.46}$ & $64.66_{\scriptstyle \pm 0.81}$ & $65.39_{\scriptstyle \pm 0.58}$ & $66.02_{\scriptstyle \pm 0.26}$ &  &  &  &  & $67.30_{\scriptstyle \pm 0.58}$ \\  
\specialrule{1.5pt}{1.5pt}{1.5pt}
\end{tabular}}

\label{tab:results}
\end{table*}

\paragraph{Conclusion}
The bound on \( \Delta_j \) demonstrates that the influence of a single node \( v_j \) in the original graph on the synthetic graph embeddings diminishes as the size of the original graph increases. This theoretical derivation validates the intrinsic privacy-preserving nature of GC, proving that the technique itself is secure. Furthermore, as highlighted in the Sec.~\ref{sec: intro}, GC transitions the potential privacy exposure from the level of the entire graph's original nodes to that of the compressed synthetic nodes, effectively reducing the granularity of sensitive information leakage. It is important to note that the provided proof primarily applies to scenarios where nodes do not overlap across different subgraphs or communities. In cases where nodes belong to overlapping clusters, additional considerations and methods may be necessary to ensure privacy protection. These results establish GC as a secure solution for privacy-sensitive FGL scenarios within the defined application boundaries.

\begin{figure}[t]
    \centering
    \subfigure[Ogbn-arxiv]{
        \includegraphics[width=0.225\textwidth]{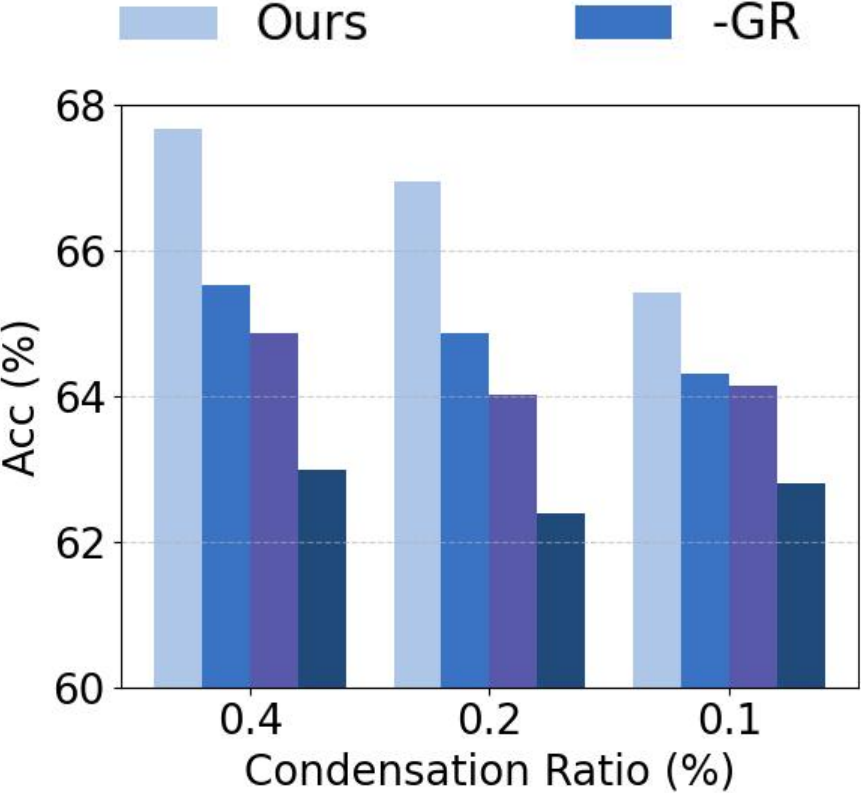}
    }
    \subfigure[Flickr]{
        \includegraphics[width=0.225\textwidth]{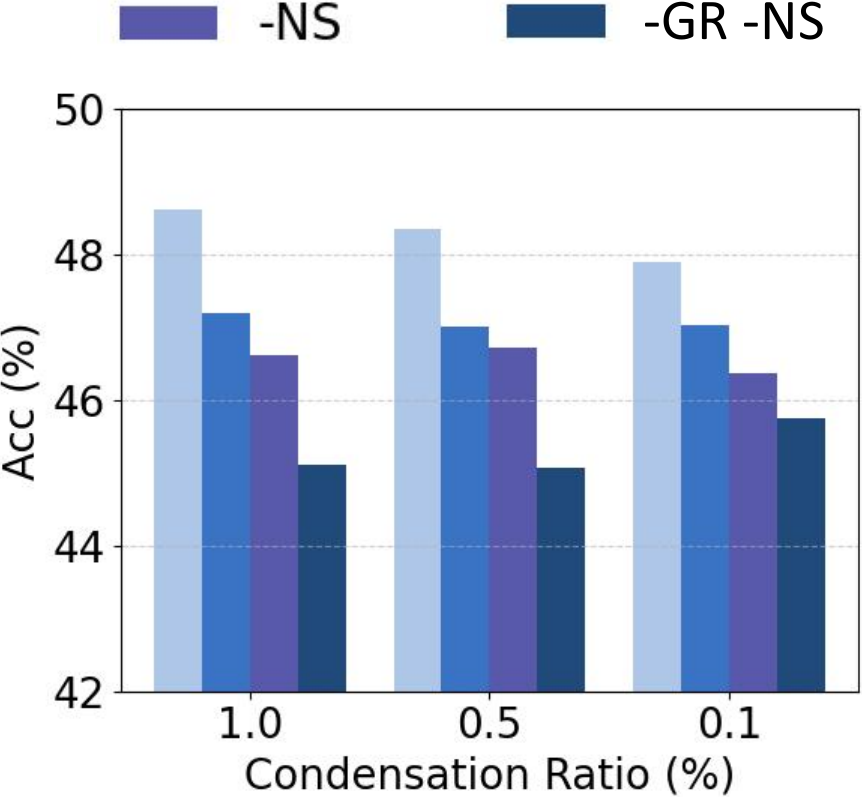}
    }
    
    \caption{Ablation Study of FedC4 on NS and GR Modules.}
    \label{fig:ablation_study}
\end{figure}

\subsection{Communication Cost}

We analyze the theoretical communication cost of FedC4 and compare it with that of the traditional S-C, personalized S-C and traditional C-C paradigms, as shown in Table~\ref{tab:communication_costs}.

\vspace{-1em}
\begin{table}[H]
\centering
\caption{Comparison of Communication Costs}
\begin{tabular}{c c}
\toprule
\textbf{Method} & \textbf{Communication Cost} \\
\midrule
S-C Paradigm (lv.1, lv.2) & $O(2 C p)$ \\
C-C Paradigm (lv.3)& $O(C^2 N d)$ \\
FedC4 (lv.4) & $O(C \log C \, N' d)$ \\
\bottomrule
\end{tabular}

\label{tab:communication_costs}
\end{table}
\vspace{-1em}

Here, $C$ is the number of clients, $p$ is the model size, $N$ is the number of nodes, $N'$ is the number of condensed nodes ($N' \ll N$), and $d$ is the embedding dimension. Traditional S-C scales linearly with $C$, C-C grows quadratically, while FedC4 reduces communication to $O(C \log C \, N' d)$ through GC and selective broadcasting in CM, achieving better scalability. The local graph condensation process is executed entirely on each individual client, incurring absolutely no additional communication costs at all.

\begin{figure}[t]
    \centering
    \subfigure[Ablation on CM module]{
        \includegraphics[width=0.22\textwidth]{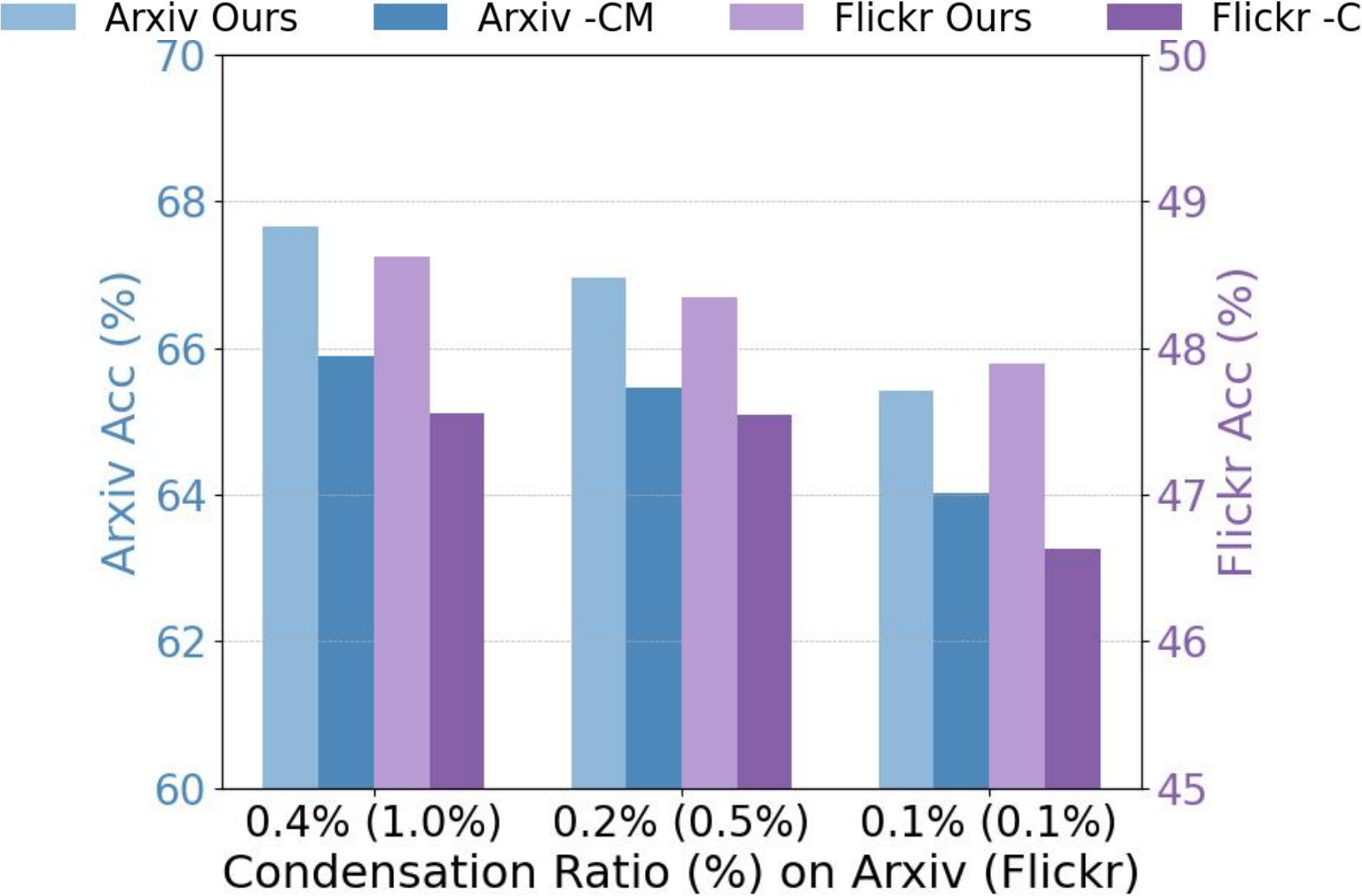}
    }
    \subfigure[Robustness Study]{
        \includegraphics[width=0.22\textwidth]{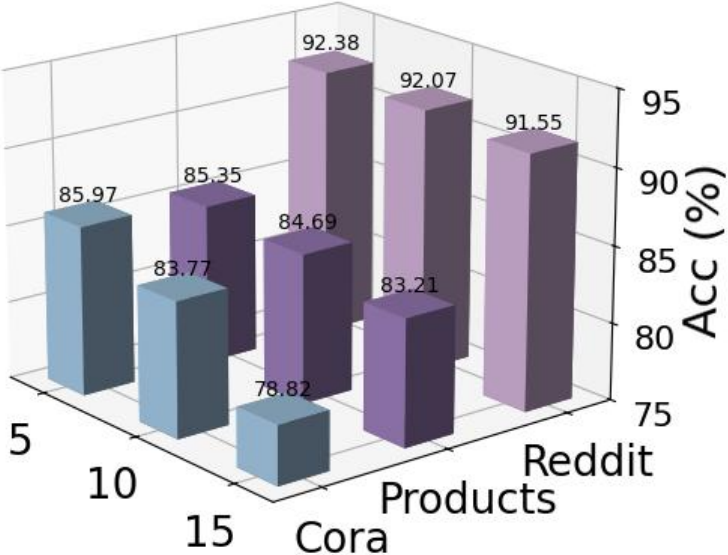}
    }
    \caption{(a) Ablation Study of FedC4 on CM Module. (b) Accuracy Comparison with varying clients (5, 10, and 15) on Cora (8\%), Products (0.4\%), and Reddit (1\%).}
    \label{fig:ablation_study_cm}
\end{figure}

\section{Experiments}
In this section, we evaluate the effectiveness of our framework, FedC4, using 8 graph datasets. We introduce baseline GC and FGL approaches, followed by a detailed experimental setup. The study addresses the following questions: \textbf{Q1:} Does FedC4 outperform state-of-the-art FGL frameworks and GC methods across diverse graph settings? \textbf{Q2:} What drives the performance improvements in FedC4? \textbf{Q3:} How does FedC4 perform in terms of efficiency and scalability compared to other FGL methods? \textbf{Q4:} Does FedC4 ensure privacy within the C-C paradigm in FGL scenarios?

\subsection{Experimental Setup}
\noindent \paragraph{Datasets} We evaluate the performance of the proposed framework on 8 datasets, including transductive and inductive settings, as well as large-scale and heterogeneous graphs. For the transductive setting, we use \textit{Cora}, \textit{Citeseer}~\cite{cora}, \textit{Ogbn-arxiv}~\cite{arxiv}, and \textit{Physics}~\cite{products&physics} datasets. For the inductive setting, we employ \textit{Flickr} and \textit{Reddit}~\cite{flickr&reddit} datasets. Additionally, we include the large-scale \textit{Ogbn-products}~\cite{products&physics} dataset and the heterogeneous \textit{Roman-empire}~\cite{roman-empire} dataset to evaluate scalability and robustness of FedC4.

\noindent \paragraph{Baselines} 
We compare our proposed method to 12 baselines: (1) two FL methods (FedAvg, FedDC)~\cite{FedAvg,feddc}, (2) three Graph Reduction methods in the FL setting (Random, Herding, Coarsening)~\cite{herding,coarsening}, (3) three GC methods in the FL settin (GCond, SFGC, SGDD)g~\cite{GCOND,SFGC,SGDD}, and (4) three C-C FGL methods (FedSage, FedDEP, FedGCN)~\cite{FedSage,feddep,fedgcn}. 
Additionally, we include the best-performing S-C FGL method (FGSSL, FedGTA, FedTAD) on each dataset~\cite{FGSSL,fedgta,fedtad}. For Graph Reduction and GC baselines, we apply client-side condensation with three different ratios and train local models on the condensed graphs. The compression ratios are determined based on prior studies~\cite{GCOND} and specific FGL settings.

\noindent \paragraph{Implementation Details} 
We simulate the subgraph systems using the Louvain community detection algorithm, specifically partitioning the original graph into 5 distinct communities. For local GNNs, we employ a 2-layer GCN with a hidden dimension of 64. All networks are optimized using SGD~\cite{SGD} with a weight decay of \(5 \times 10^{-4}\). 
Each client is trained for 5 local epochs with 200 communication rounds. 
For the parameters of the NS and GR module, we set \(\tau = 0.38\), \(\alpha = 150\), and \(\beta = 250\).

\begin{figure}[t]
    \centering
    \subfigure[Node Selector]{
        \includegraphics[width=0.22\textwidth]{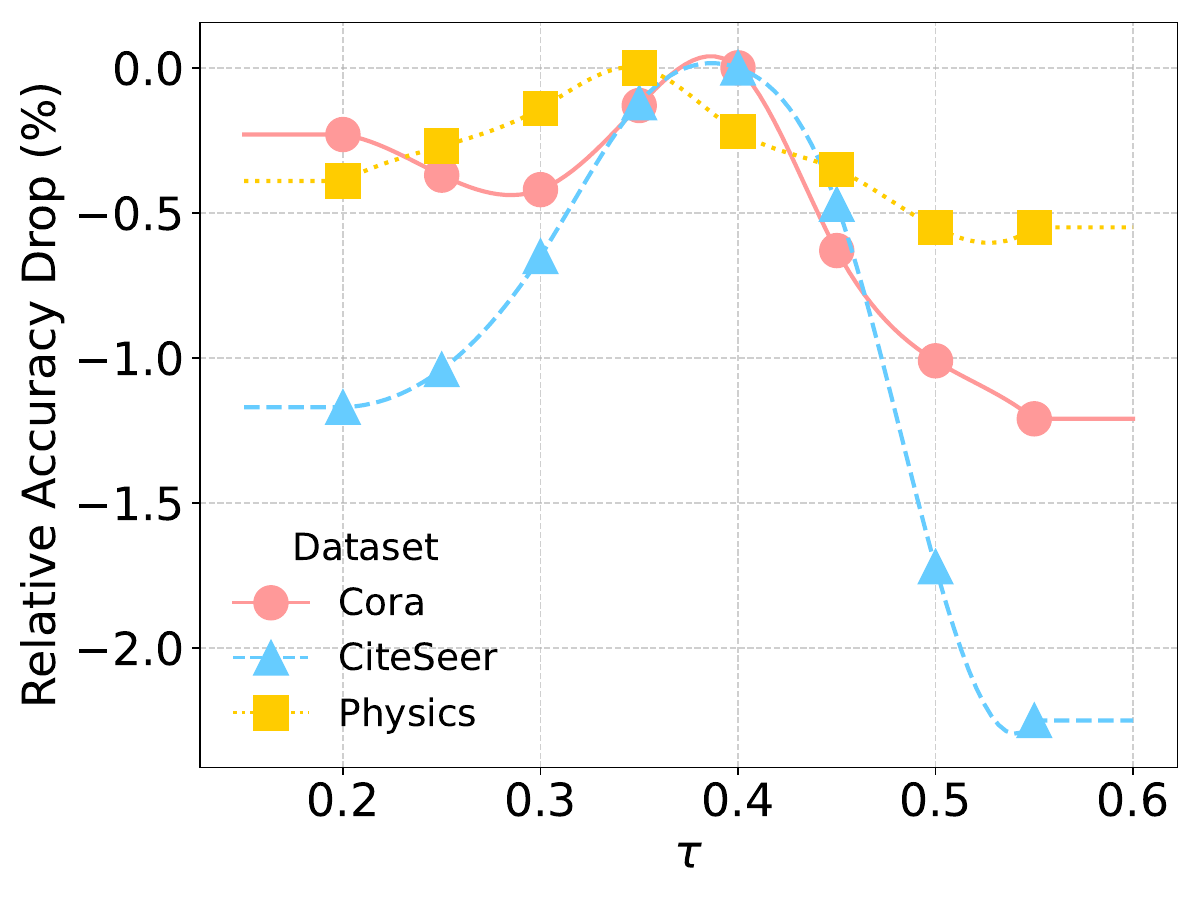}
    }
    \subfigure[Graph Rebuilder]{
        \includegraphics[width=0.22\textwidth]{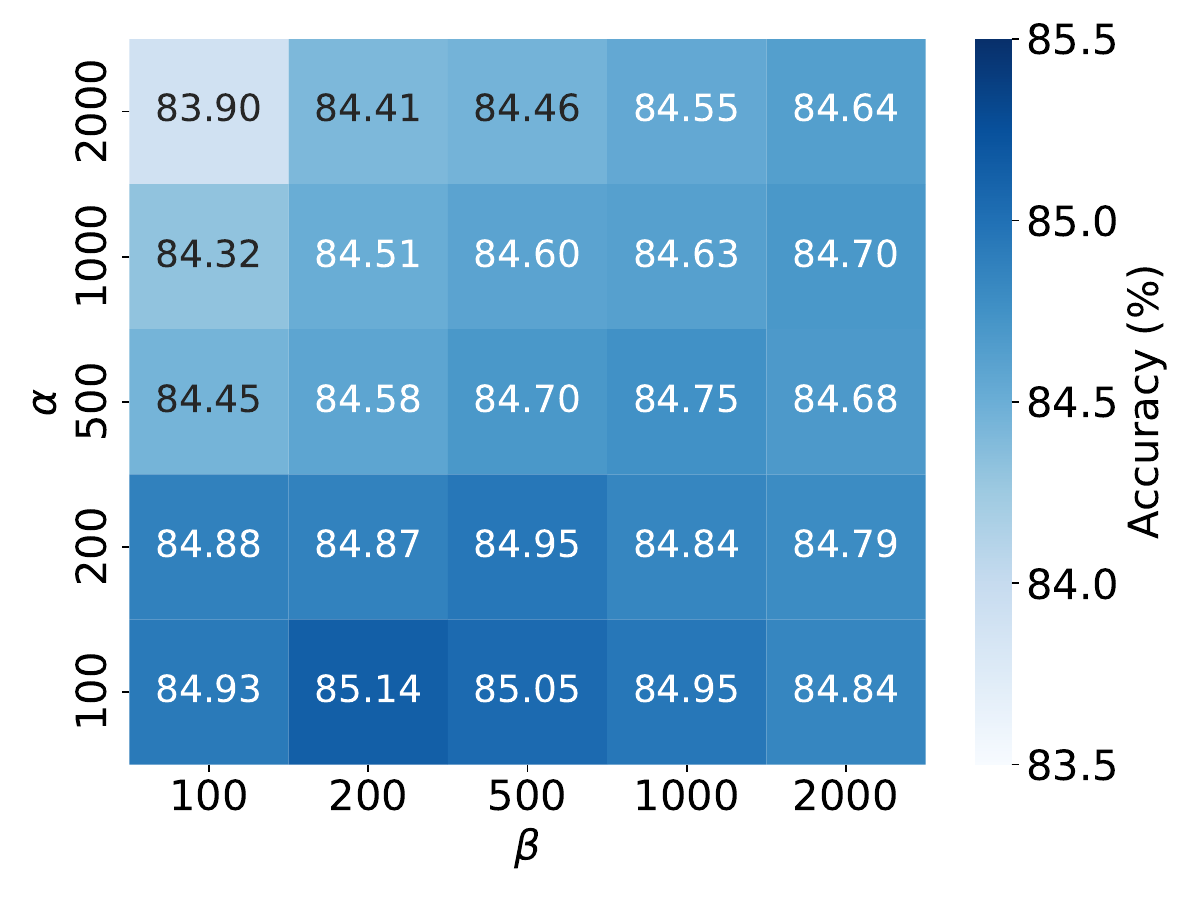}
    }
    \caption{Hyper-parameter study for (a) $\tau$ in Node Selector Module. (b) $\alpha$ , $\beta$ in Graph Rebuilder Module.}
    \label{fig:hyper_study}
\end{figure}

\begin{figure}[t]
    \centering
    \includegraphics[width=0.45\textwidth]
        {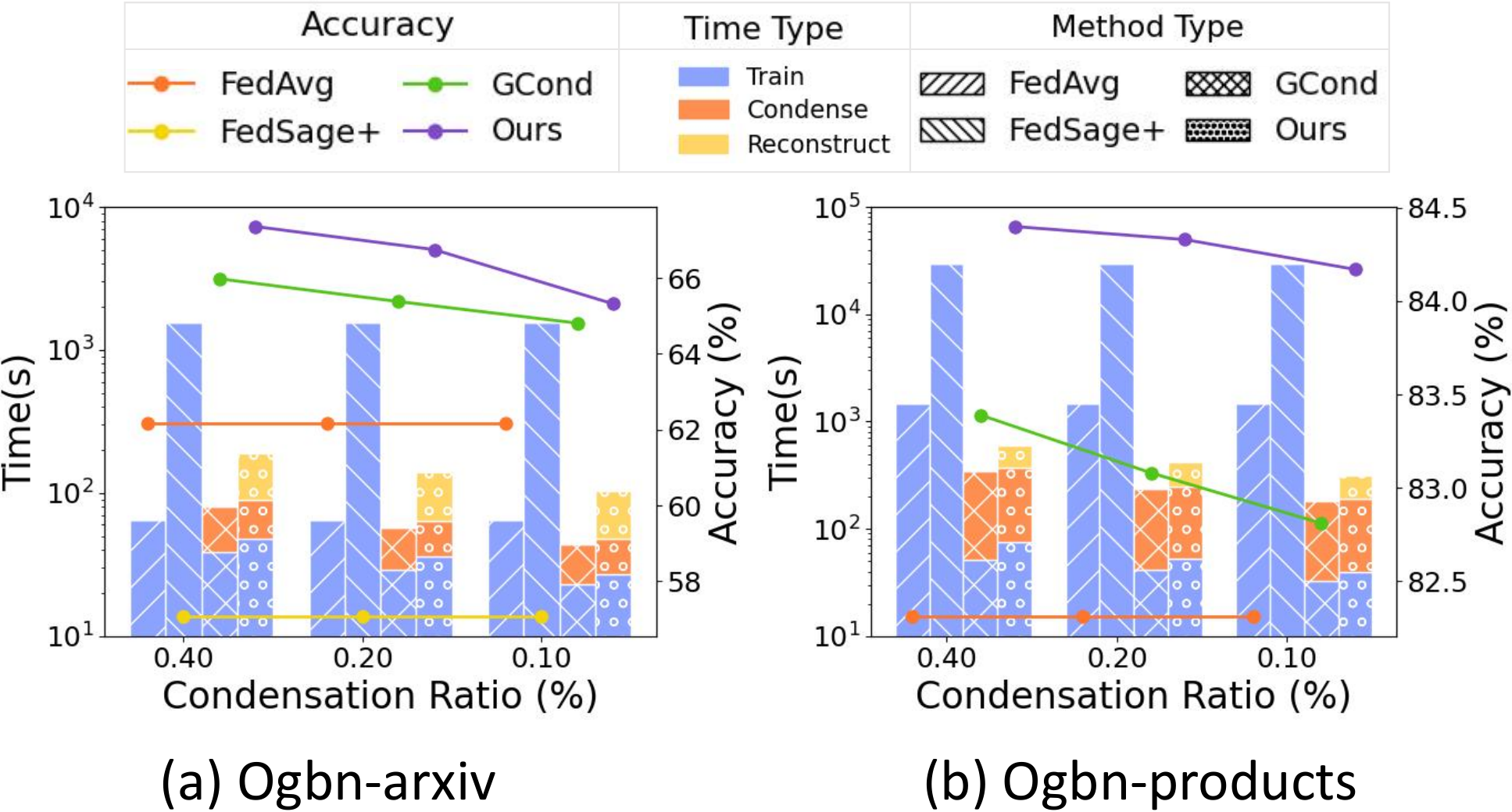}
     
    \caption{Efficiency evaluation on Arxiv and Products.}
    \label{fig:efficiency}
\end{figure}

\begin{figure}[t]
    \centering
    \subfigure[Convergence Curves]{
        \includegraphics[width=0.22\textwidth]{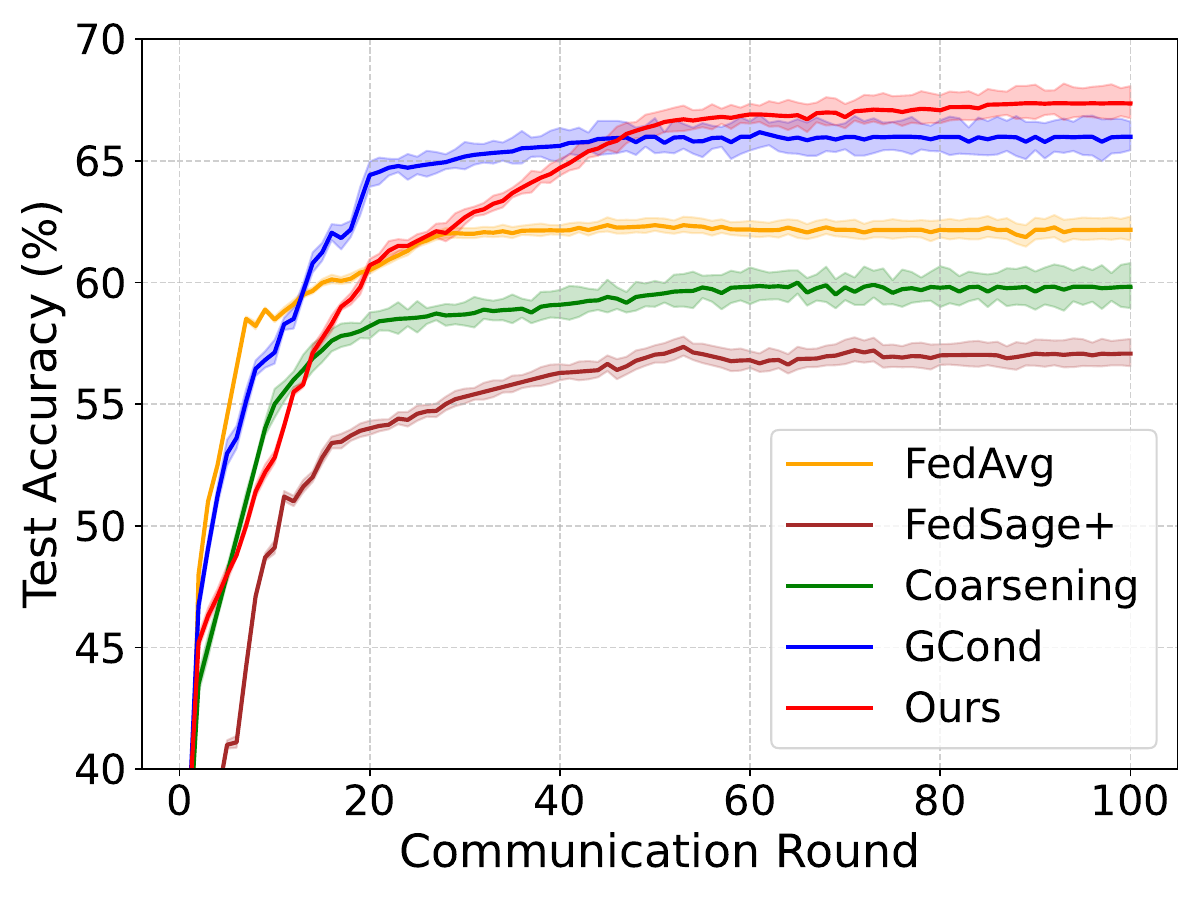}
    }
    \subfigure[Impact of Privacy Scale]{
        \includegraphics[width=0.22\textwidth]{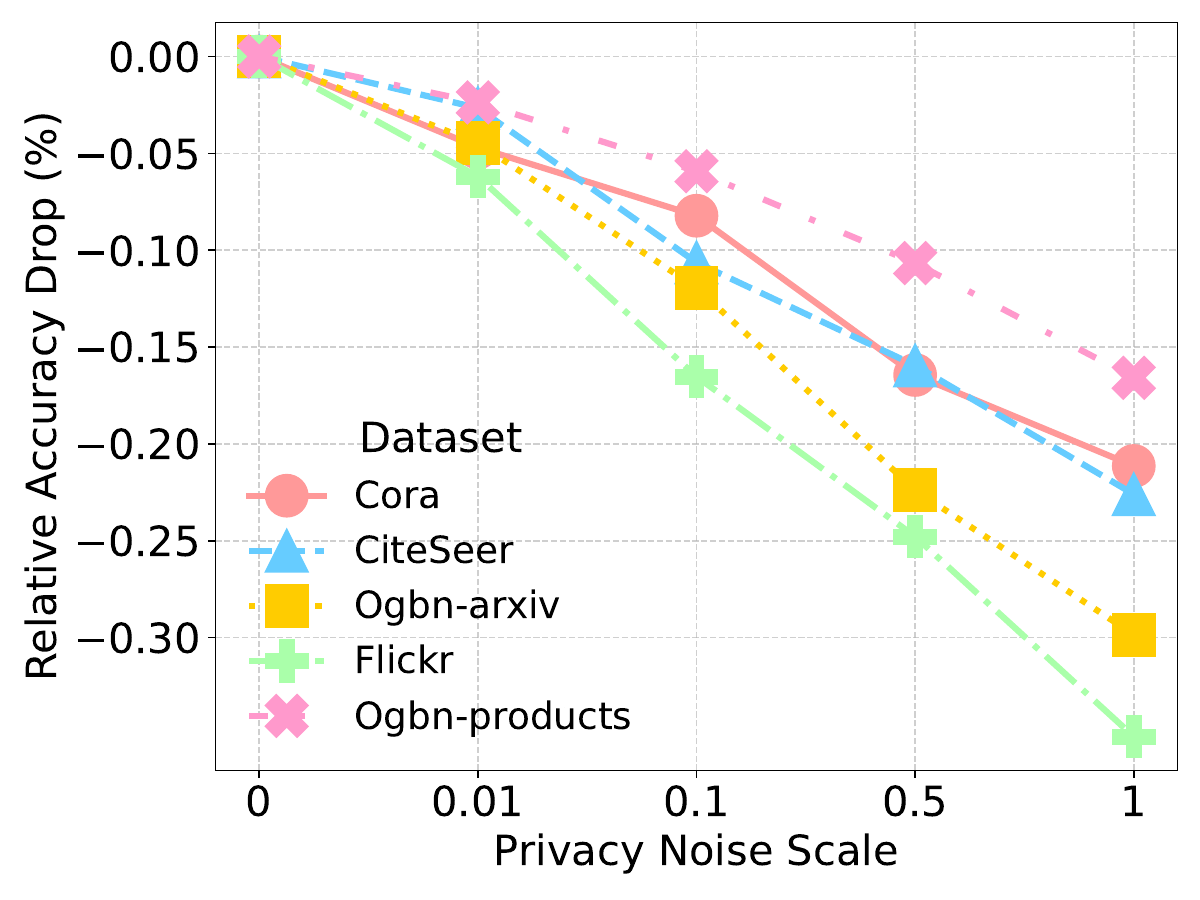}
    }
    \caption{(a) Convergence curves. 
             (b) Impact of privacy scale.}
    \label{fig:traincurve}
\end{figure}

\subsection{Performance Comparison}
\label{par:performance}
To answer \textbf{Q1}, the results in Table~\ref{tab:results} demonstrate that FedC4 consistently outperforms most baseline methods across FL, FL+Graph Reduction, FL+GC, and FGL categories: \textbf{(1) FL methods}: Compared to FedAvg and FedDC, FedC4 achieves an average improvement of 12.03\% by leveraging the structural information of graph data, effectively addressing data heterogeneity and overcoming the limitations of vanilla aggregation. \textbf{(2) FL+Graph Reduction methods}: Compared to Herding and Coarsening, FedC4 not only maintains efficiency but also improves accuracy significantly by 7.88\% and 7.76\%. GC strategies consistently outperform other graph reduction techniques in both accuracy and scalability, making GC the preferred choice for our framework. \textbf{(3) FL+GC methods}: FL+GC methods such as GCond and SFGC, while achieving efficiency gains, suffer from inter-client heterogeneity. Compared to GCond, FedC4 achieves an accuracy improvement of 3.01\%, and compared to SFGC, FedC4 improves by 2.69\%. \textbf{(4) FGL methods}: In the FGL category, specialized methods like FedSage+ (C-C) and FGSSL (S-C) face scalability challenges, often resulting in OOM or OOT errors on large-scale datasets. Compared to FGL methods, FedC4 improves by an average of 7.52\% over S-C methods and 3.58\% over other C-C methods. These performance comparisons between S-C, C-C, and FedC4 validate the communication granularity issues in Sec.~\ref{sec: intro}, with FedC4 overcoming the limitations of traditional S-C and C-C methods through more fine-grained personalized communication.

\subsection{Ablation Study}
To answer \textbf{Q2}, we conduct an ablation study by systematically removing Graph Rebuilder (-GR) and Node Selector (-NS) modules. Ogbn-arxiv, a moderate-size transductive dataset, and Flickr, an inductive dataset, were chosen for the ablation study to evaluate the performance of the modules in different learning settings. As shown in Fig.~\ref{fig:ablation_study}, combining all modules achieves the highest performance, validating the effectiveness of fine-grained communication. 

In addition to the GR and NS modules, we also conducted an ablation study on the CM module to explore the impact of different broadcasting strategies on model performance. Specifically, we compared the effects of full broadcasting and selective broadcasting within the CM module. In terms of theoretical communication cost, full broadcasting incurs a cost of \(O(C^2 N' d)\), whereas selective broadcasting significantly reduces this cost to \(O(C \log C\, N' d)\). As shown in Fig.~\ref{fig:ablation_study_cm}(a), selective broadcasting within the CM module achieves superior performance by reducing unnecessary communications and focusing on the most pertinent cross-client information exchanges. This result further confirms the effectiveness of fine-grained communication strategies in simultaneously enhancing performance and communication efficiency.

In addition to performance-level ablations, we also evaluate the structural impact of the GR module to understand how well it recovers the original graph topology after condensation in Appendix~\ref{sec:gr_analysis}.

\subsection{Efficiency and Scalability Study}
To answer \textbf{Q3}, we conduct four experiments: a Robustness Study to investigate the impact of client number variation on model performance, an Efficiency Study to evaluate efficiency and accuracy, a Hyper-parameter Study to analyze the impact of key parameters in NS and GR module, and a Convergence Study to compare the training stability of FedC4 with other methods.

\noindent \paragraph{Robustness study}
To assess the robustness of our method to different client configurations, we conduct a study under varying numbers of clients (5, 10, and 15) on three datasets: Cora (8\%), Products (0.4\%), and Reddit (1\%). As shown in Fig.~\ref{fig:ablation_study_cm}(b), FedC4 consistently maintains high accuracy as the number of clients increases, indicating strong robustness to data partitioning granularity. This robustness study highlights the stability of our method across different federated settings (number of clients).

\noindent \paragraph{Efficiency study}
To evaluate the efficiency of our proposed method, we conduct experiments on two datasets, Arxiv and Products, under varying condensation ratios (0.40\%, 0.20\%, 0.10\%). As shown in Fig.~\ref{fig:efficiency}(a) and (b), FedC4 achieves superior accuracy compared to baseline approaches (FedAvg, FedSage+ and GCond), while maintaining a competitive efficiency. (FedSage+ OOM on products)

Although the condensation and graph reconstruction processes introduce additional computation, they significantly reduce the overall training time by producing more compact and informative graph representations. This trade-off effectively speeds up model training, highlighting the practicality of integrating GC into C-C FGL for efficiency and scalability.

\noindent \paragraph{Hyper-parameter study} 
\label{par:hyper}
We analyze the performance under different values of $\tau$ (Node Selector) and $\alpha, \beta$ (Graph Rebuilder). These parameters control node selection and graph reconstruction during federated learning. Results are shown in Fig.~\ref{fig:hyper_study}, where Fig.~\ref{fig:hyper_study}(a) illustrates the impact of varying $\alpha$ and $\beta$ with a fixed $\tau$, highlighting how different feature weightings affect model performance. Fig.~\ref{fig:hyper_study}(b) shows the effect of varying $\tau$ while fixing $\alpha$ and $\beta$, indicating how NS module influences model accuracy.

\noindent \paragraph{Convergence study} 
Fig.~\ref{fig:traincurve}(a) shows the average test accuracy curves during training across five random runs on Ogbn-arxiv. It can be observed that FedC4 consistently outperforms baselines in terms of average test accuracy and achieves faster convergence. It demonstrate that FedC4 not only performs well but also stabilizes quickly, highlighting its robustness and efficiency in FGL settings.

\subsection{Privacy Protection}
To answer \textbf{Q4}, we refer to the theoretical analysis in Privacy for Free\cite{privacy}, which proves that propagating condensed datasets is inherently secure. In Sec.~\ref{sec: privacy}, we extend this proof to graph data, demonstrating that transmitting node embeddings of condensed graphs also satisfies privacy guarantees.

Building on this theoretical foundation, we introduced varying levels of Laplace noise during the condensation process to simulate real-world privacy challenges. As shown in Fig.~\ref{fig:traincurve}(b), results on real-world datasets show that FedC4 maintains high accuracy even under significant noise, confirming its robustness in privacy-preserving FGL and demonstrating its effectiveness in maintaining model performance while safeguarding privacy.

\section{Conclusion}
This paper presents the first integration of GC into the C-C paradigm in FGL, introducing three novel modules to address key challenges: fine-grained personalization, communication overhead, and privacy risks. The fine-grained personalization challenge, stemming from the need for tailored information exchange, is effectively addressed through three core modules: CM selectively shares global statistics tailored to client-specific contexts, NS identifies representative nodes for targeted knowledge transfer, and GR reconstructs personalized graph structures, enabling efficient and fine-grained collaboration. To tackle both communication overhead and privacy concerns, GC techniques are seamlessly integrated. Theoretical analysis further validates that GC significantly reduces communication costs while mitigating privacy risks. Experimental results demonstrate that FedC4 achieves superior performance and efficiency compared to competitive baselines.

\appendix
\section*{Appendix}
\addcontentsline{toc}{section}{Appendix}

\section{Algorithm}

\subsection{Customizer Module}
\vspace{-1em}

\begin{algorithm}
\caption{Customizer Module}
\label{alg:broadcaster}
\raggedright
\textbf{Input}: Local node embeddings $\mathbf{H}_c = \{ \mathbf{h}_i \mid i \in \mathcal{V}_c \}$ \\

\begin{algorithmic}[1]
    \FOR{each communication round $t = 1, \dots, T$}
        \FOR{each client $c$}
            \STATE Compute statistics \& Normalization:
            Eq.~\ref{eq:cm8}, Eq.\ref{eq:cm9}, Eq.\ref{eq:cm10}
            \IF{$t = 1$} 
                \STATE Broadcast statistics to \textbf{all clients}
            \ELSE
                \STATE Use the clustering results from the previous round's Node Selector: $\mathcal{C}_{\text{same}}$
                \STATE Broadcast statistics to clients in $\mathcal{C}_{\text{same}}$: Eq.~\ref{eq:cm11}
            \ENDIF
        \ENDFOR
    \ENDFOR
\end{algorithmic}
\end{algorithm}
\vspace{-2.5em}
\subsection{Node Selector Module}
\vspace{-1em}

\begin{algorithm}
\caption{Node Selector Module}
\label{alg:nodeselector}
\raggedright
\textbf{Input}: Embedding distributions $\{\text{Dis}_c \mid c \in \mathcal{C}\}$, mean embeddings $\{\mu_c \mid c \in \mathcal{C}\}$\\

\begin{algorithmic}[1]
    \FOR{$c \in \mathcal{C}$}
        \STATE Compute Sliced Wasserstein Distances: Eq.~\ref{SWD}
        \STATE Cluster clients:
        \[
        \mathcal{C}_c = \left\{ c' \mid \text{SWD}_{c,c'} \leq \delta \right\}, 
        \quad \forall c \in \mathcal{C}
        \]
        \FOR{$c' \in \mathcal{C}_c$}
            \STATE Compute similarity: Eq.~\ref{cossim}
            \STATE Select nodes:
            \[
            \mathcal{S}_c = \left\{ \mathbf{h}_i \mid S(\mathbf{h}_i, \mu_{c'}) > \tau, 
            \quad \forall c' \in \mathcal{C}_c \right\}
            \]
        \ENDFOR
    \ENDFOR
    \STATE \textbf{Return} $\{\mathcal{S}_c \mid c \in \mathcal{C}\}$
\end{algorithmic}
\end{algorithm}

\subsection{Graph Rebuilder Module}
The Graph Rebuilder module reconstructs reliable graph structures for selected nodes to preserve connectivity and information flow after graph condensation. By adaptively restoring topology based on node relationships, it mitigates structural loss and helps the model capture key dependencies. Algorithm~\ref{alg:graph_rebuilder} outlines this process.

\begin{algorithm}
\caption{Graph Rebuilder Module}
\label{alg:graph_rebuilder}
\raggedright
\textbf{Input}: Node features $\{\mathbf{X_c}\}$ , selected nodes $\{\mathbf{S_c}\}$ \\
\textbf{Output}: Rebuilt graph $\mathbf{G_c}$ \\
\begin{algorithmic}[1]
    \FOR{each node $i$ in $\{\mathbf{S_c}\}$}
        \FOR{each neighbor $j$ of node $i$}
            \STATE Compute similarity: Eq.\ref{embsim}
        \ENDFOR
    \ENDFOR
    \STATE Compute the reconstruction loss: Eq.\ref{lossrec}
    \STATE Optimize $\mathbf{Z}$ to minimize $\mathcal{L}_{\text{rec}}$
    \STATE \textbf{Return} Rebuilt graph $\mathbf{G_c} = \{\mathbf{Z}, \mathbf{X}_c\}$
\end{algorithmic}
\end{algorithm}

\section{Discussion of GR Module}
\label{sec:gr_analysis}
To assess the GR module’s ability to restore topological information after condensation, we analyze three structural metrics: degree distribution KL divergence, graph density, and homophily.

\begin{table}[htbp]
\centering
\caption{Topological metrics comparison among original, condensed, and reconstructed graphs.}
\label{tab:topology_analysis}
\begin{tabular}{lccc}
\toprule
\textbf{Metric} & \textbf{Original} & \textbf{Condensed} & \textbf{Reconstructed} \\
\midrule
KL Divergence ↓ & 0.000 & 0.612 & 0.105 \\
Graph Density ↓   & 0.007 & 0.855 & 0.059 \\
Homophily ↑   & 0.731 & 0.422 & 0.694 \\
\bottomrule
\end{tabular}
\end{table}

As shown in Table~\ref{tab:topology_analysis}, condensation leads to significant structural deviation. The KL divergence rises to 0.612, indicating notable changes in degree distribution. Graph density increases drastically from 0.007 to 0.855, suggesting an overly connected structure, while homophily drops from 0.731 to 0.422, weakening semantic consistency between connected nodes.

In contrast, the reconstructed graph significantly reduces divergence across all metrics: KL divergence decreases to 0.105, density is corrected to 0.059, and homophily rises to 0.694. These improvements show that GR effectively recovers both sparse connectivity and meaningful label relationships. Notably, although homophily is often emphasized in homophilic datasets, GR also proves effective on heterophilic graphs. For example, our accuracy on the \textit{Roman-empire} dataset in Sec.~\ref{par:performance} confirm GR’s adaptability under heterophily conditions.

\section{Training Configuration}
\subsection{Statistics of Datasets}

We evaluate our proposed method on eight publicly available graph datasets widely used in federated and decentralized learning research, including citation networks (Cora, Citeseer, Ogbn-arxiv), co-purchase networks (Ogbn-products), social and community graphs (Reddit, Flickr), physical systems (Physics), and heterophilic graphs (Roman-empire). The datasets vary significantly in terms of scale, feature dimensionality, and class distributions, thus providing a comprehensive benchmark for assessing the scalability, generalizability, and robustness of our approach. Each dataset is partitioned into training, validation, and testing subsets. Specifically, we use a 60\%/20\%/20\% split for Cora, Citeseer, Arxiv, Physics, and Reddit, a 50\%/25\%/25\% split for Flickr and Empire, and a 30\%/10\%/60\% split for Products. Detailed statistics are summarized in Table~\ref{tab:dataset_statistics}.

\begin{table}[H]
\centering
\caption{Statistics of Datasets.}
\begin{tabular}{lrrrr}
\specialrule{1.5pt}{1.5pt}{1.5pt}
\textbf{Dataset}       & \textbf{Nodes} & \textbf{Edges} & \textbf{Features} & \textbf{Classes} \\ 
\midrule
Cora                  & 2,708          & 5,429          & 1,433             & 7                \\ 
Citeseer              & 3,327          & 4,732          & 3,703             & 6                \\ 
Arxiv                 & 169,343        & 1,166,243      & 128               & 40               \\ 
Physics               & 34,493         & 495,924        & 841               & 5                \\ 
Flickr                & 89,250         & 899,756        & 500               & 7                \\ 
Reddit                & 232,965        & 11,606,919     & 602               & 41               \\ 
Products              & 2,449,029      & 61,859,140     & 100               & 47               \\ 
Empire                & 22,678         & 342,334        & 50                & 18               \\ 
\specialrule{1.5pt}{1.5pt}{1.5pt}
\end{tabular}
\label{tab:dataset_statistics}
\end{table}

\subsection{Computation Resource}
Experiments are conducted on a Linux server equipped with 2
Intel(R) Xeon(R) Gold 6240 CPUs @ 2.60GHz (36 cores per socket,
72 threads total), 251GB RAM (with approximately 216GB available),
and 4 NVIDIA A100 GPUs with 40GB memory each. The software
environment includes Python 3.11.10 and PyTorch 2.1.0 with CUDA
11.8. We adopt a multi-GPU training setup where each simulated
client is assigned to a dedicated GPU during federated optimization.
This design enables parallel execution across clients in the Client-Client (C-C) communication paradigm.

The 4 NVIDIA A100 GPUs provide substantial computational
power and memory capacity, enabling efficient handling of large-scale datasets such as Ogbn-products. However, during experiments,
OOM (Out of Memory) occurs when a model exceeds the 40GB
memory capacity of a single A100 GPU, while OOT (Out of Time)
is defined as operations exceeding 8 hours of computation time on
a single A100 GPU. These constraints highlight the challenges in
scaling graph learning tasks under limited hardware resources.

In contrast to existing FGL methods that frequently suffer from
OOM or OOT issues on large-scale graphs, FedC4 demonstrates superior scalability and efficiency. By employing graph condensation
and selective communication, FedC4 significantly reduces the memory footprint and computational load during training. This allows
it to complete training on massive datasets such as Ogbn-products
without encountering resource bottlenecks, thereby validating its
practicality and robustness in real-world federated settings.

\section*{Acknowledgments}
\addcontentsline{toc}{section}{Acknowledgment}
This work is supported by the NSFC Grants U2241211, 62427808 and U24A20255. Rong-Hua Li is the corresponding author of this paper.

\section*{GenAI Usage Disclosure}
\addcontentsline{toc}{section}{GenAI Usage Disclosure}
This article utilized ChatGPT-4o and Grok-3 for textual improvement and experimental code restructuring.


\bibliographystyle{ACM-Reference-Format}
\balance
\bibliography{references}


\end{document}